\pdfoutput=1

\documentclass[11pt]{article}

\usepackage[preprint]{acl}

\usepackage{times}
\usepackage{latexsym}
\usepackage{amssymb}
\usepackage{amsmath}
\usepackage{amsthm}
\usepackage{multirow}
\usepackage{array}
\usepackage{booktabs}
\usepackage{makecell}
\usepackage{url}
\newtheorem{theorem}{Theorem}
\newtheorem*{theorem*}{Theorem}

\usepackage[T1]{fontenc}

\usepackage[utf8]{inputenc}

\usepackage{microtype}

\usepackage{inconsolata}

\usepackage{graphicx}

\title{\emph{AlignDistil}: Token-Level Language Model Alignment \\as Adaptive Policy Distillation}

\author{Songming Zhang\textsuperscript{1,2}\thanks{\ Work was done when Songming was interning at Tencent.}, Xue Zhang\textsuperscript{1,2}, Tong Zhang\textsuperscript{3}, Bojie Hu\textsuperscript{3}, \\
\textbf{Yufeng Chen}\textsuperscript{1,2}\thanks{ \ Yufeng Chen is the corresponding author.}, and 
\textbf{Jinan Xu}\textsuperscript{1,2} \\
\textsuperscript{1}Key Laboratory of Big Data \& Artificial Intelligence in Transportation,\\(Beijing Jiaotong University), Ministry of Education \\
\textsuperscript{2}School of Computer Science and Technology, Beijing Jiaotong University, Beijing, China \\
\textsuperscript{3} Tencent Inc, China \\
\texttt{\{smzhang22,zhang\_xue,chenyf,jaxu\}@bjtu.edu.cn}}

\begin{document}
\maketitle
\begin{abstract}
In modern large language models (LLMs), LLM alignment is of crucial importance and is typically achieved through methods such as reinforcement learning from human feedback (RLHF) and direct preference optimization (DPO).
%
%
%
However, in most existing methods for LLM alignment, all tokens in the response are optimized using a sparse, response-level reward or preference annotation.
The ignorance of token-level rewards may erroneously punish high-quality tokens or encourage low-quality tokens, resulting in suboptimal performance and slow convergence speed.
To address this issue, we propose \emph{\textbf{AlignDistil}}, an RLHF-equivalent distillation method for token-level reward optimization.
Specifically, we introduce the reward learned by DPO into the RLHF objective and theoretically prove the equivalence between this objective and a token-level distillation process, where the teacher distribution linearly combines the logits from the DPO model and a reference model.
On this basis, we further bridge the accuracy gap between the reward from the DPO model and the pure reward model, by building a contrastive DPO reward with a normal and a reverse DPO model.
Moreover, to avoid under- and over-optimization on different tokens, we design a token adaptive logit extrapolation mechanism to construct an appropriate teacher distribution for each token.
%
%
Experimental results demonstrate the superiority of our AlignDistil over existing methods and showcase fast convergence due to its token-level distributional reward optimization.

\end{abstract}

\begin{figure}
    \centering
    \includegraphics[width=\linewidth]{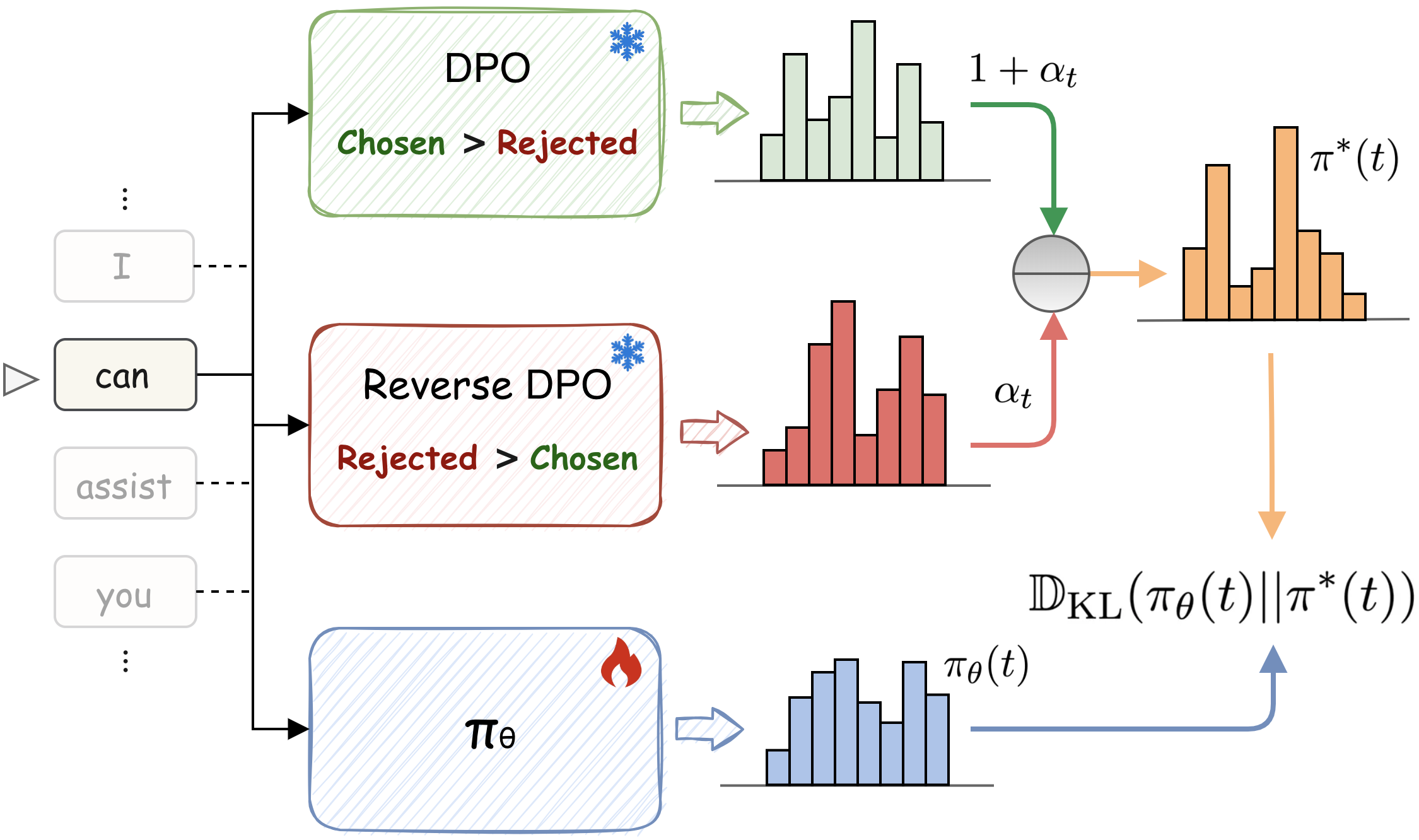}
    \caption{An overview of our AlignDistil. At token position $t$, the distribution from the current policy $\pi_{\theta}(t)$ is guided by a teacher distribution $\pi^{*}(t)$, which is constructed from an adaptive extrapolation between logit distributions from a DPO model and a reverse DPO model with a weight $\alpha_t$.}
    \label{fig:method}
\end{figure}

\section{Introduction}
Current large language models (LLMs) have demonstrated remarkable capabilities in producing human-desired outputs under different circumstances \cite{bai2022anthropic_rlhf,ouyang2022instructgpt,meta2024llama3}.
This is largely achieved by a key procedure in the post-training of LLMs, \emph{i.e.}, LLM alignment with human preference.
Existing solutions for LLM alignment mainly includes reinforcement learning from human feedback (RLHF) \cite{christiano2017deeprlhf,stiennon2020summarize_rlhf,bai2022anthropic_rlhf,ouyang2022instructgpt} and direct preference learning algorithms \citep{rafailov2024dpo,azar2024ipo,ethayarajh2024kto}.
Therein, RLHF is a two-stage method that first 1) trains a response-level reward model based on human preference labels, and then 2) optimizes the policy model with RL algorithms under this reward model while preventing deviation from the initial model.
%
%
Alternatively, direct preference learning algorithms, \emph{e.g.}, direct preference optimization (DPO, \citealt{rafailov2024dpo}), simplify RLHF via parameterizing the reward with the policy model and directly training it on the preference data.
%

%
Despite their prevalence, most existing methods for LLM alignment optimize tokens with a sparse, response-level reward or preference annotation.
However, this response-level feedback is coarse-grained and lacks reflection on the individual contribution of each token in the response \citep{yoon2024tlcr,li2024rltokenfb,xia2024inverseq*,yang2024sepo}, which may erroneously punish tokens with high quality or encourage tokens with low quality.
Consequently, those methods based on response-level feedback have been revealed with limitations on both performance and convergence speed \citep{chan2024densereward,zhong2024rto,liu2024tisdpo}.

To address this issue, in this paper, we propose \textbf{\textit{AlignDistil}} (as shown in Figure \ref{fig:method}), a simple distillation method derived from the RLHF objective for token-level reward optimization.
Specifically, our method starts from introducing the DPO reward \citep{rafailov2024dpo} into the original objective of RLHF.
Based on the property of token-level decomposition of the DPO reward \citep{rafailov2024fromrtoq}, we prove a theoretical equivalence between the original sequence-level objective of RLHF and a token-level distillation objective.
In this distillation objective, the current policy is guided by a teacher distribution that linearly combines the logit distribution output from the two LLMs in the DPO reward.
Built on this theoretical finding, our AlignDistil further involves two targeted designs for token-level optimization.
Firstly, given that rewards from DPO generally perform worse than those from pure reward models \citep{lin2024limitdporeward}, we use a contrastive DPO reward for AlignDistil with a DPO model and a reverse DPO model \citep{liu2024tisdpo}, which yields better generalization performance than the vanilla DPO reward.
Furthermore, to mitigate imbalanced under- and over-optimization across different tokens, we design a token adaptive logit extrapolation mechanism to construct an appropriate teacher distribution for each token position.
%
%
Overall, our AlignDistil uses a simple distillation objective to achieve token-level reward optimization.
Additionally, its training can flexibly switch between on-policy and off-policy, trading off between effectiveness and efficiency.

We evaluate the effectiveness of our method on three common benchmarks for LLM alignment, \emph{i.e.}, AlpacaEval 2.0 \citep{dubois2024alpacaeval}, MT-Bench \citep{zheng2023mtbench} and Arena-Hard \citep{li2024arenahard}.
Experimental results demonstrate the superiority of our AlignDistil over existing methods and showcase the effectiveness of the targeted designs in the method.
Moreover, AlignDistil exhibits a faster convergence speed compared to the variants with response-level and token-level scalar-type rewards, highlighting the advantage of token-level distributional reward optimization.

In a nutshell, the contributions of this paper are as follows:
\begin{itemize} 
    \item We build a theoretical equivalence between RLHF with DPO reward and a distillation process, which offers a new perspective for performing token-level reward optimization.
    \item On this basis, we design AlignDistil, a simple distillation method with a contrastive DPO reward and a token adaptive logit extrapolation.
    \item Experimental results showcase that AlignDistil significantly outperforms existing methods and achieves faster convergence due to the token-level distributional reward optimization.
\end{itemize}
%
%
%
%
%

\section{Preliminary}
\subsection{Reinforcement Learning from Human Feedback}
Generally, RLHF contains two stages, \emph{i.e.}, reward modeling and policy optimization.
\paragraph{Reward Modeling.} 
Reward modeling generally needs a human-labeled preference dataset with $N$ samples $D=\{(x,y_w,y_l)_{i}\}_{N}$, where $x$ is the prompt from the user, and $y_w/y_l$ represents the human-annotated preferred/dispreferred response.
Then, the human preference within the data is modeled by a reward model using the Bradley-Terry model \cite{bradley1952bradleyterry}, which optimizes the reward $r_{\phi}$ with the following loss function:
\begin{align} \label{eq:reward_modeling}
    &\mathcal{L}_{\rm RM}(\phi)= \nonumber \\
    &-\!\!\!\!\mathop{\mathbb{E}}_{(x,y_w,y_l) \sim D} \Big[\log \sigma (r_{\phi}(x,y_w) - r_{\phi}(x,y_l)) \Big].
\end{align}
\paragraph{Policy Optimization.} 
Afterward, the policy model $\pi_{\theta}$ (\emph{i.e.}, the LLM) is optimized with RL algorithms like PPO to maximize its expected reward while preventing $\pi_{\theta}$ from being too far from the reference model $\pi_{\rm ref}$:
\begin{align} \label{eq:rlhf_obj}
    &\mathcal{J}_{\rm RLHF}(\theta)= \nonumber \\
    &\max_{\theta}\mathop{\mathbb{E}}_{\substack{x \sim D \\ y \sim \pi_{\theta}(\cdot|x)}} \Big[ r_{\phi}(x,y) - \beta \log \frac{\pi_{\theta}(y|x)}{\pi_{\rm ref}(y|x)} \Big],
\end{align}
where $\beta$ is a hyper-parameter to control the Kullback-Leibler (KL) divergence \citep{kullback1951kl} from the reference model.

\subsection{Direct Preference Optimization and its Implicit Reward}
\label{sec:background_dpo}
Although RLHF is proposed as the initial solution for LLM alignment, the process is somewhat complicated and expensive.
To address this, \citet{rafailov2024dpo} propose direct preference optimization (DPO) to directly train the LLM in the reward modeling stage.
Specifically, they leverage the closed-form solution of the RLHF objective and parameterize the reward with a log ratio:
\begin{equation} \label{eq:init_dpo_reward}
    r_{\theta}(x,y)=\beta\log\frac{\pi_{\theta}(y|x)}{\pi_{\rm ref}(y|x)} + \beta\log Z(x),
\end{equation}
where $Z(x)$ is the partition function and independent to $y$.
Then the training objective of DPO is derived by substituting Eq. (\ref{eq:init_dpo_reward}) into Eq. (\ref{eq:reward_modeling}):
\begin{align} \label{eq:dpo_loss}
    &\mathcal{L}_{\rm DPO}(\theta)=-{\mathbb{E}}_{(x,y_w,y_l)\sim D}\Big[ \nonumber \\
    &\log \sigma \Big(\beta\log\frac{\pi_{\theta}(y_w|x)}{\pi_{\rm ref}(y_w|x)} - \beta\log\frac{\pi_{\theta}(y_l|x)}{\pi_{\rm ref}(y_l|x)}\Big) \Big].
\end{align}
Besides, \citet{rafailov2024dpo} point out that $Z(x)$ in Eq. (\ref{eq:init_dpo_reward}) can be omitted without loss of generality:
\begin{equation} \label{eq:dpo_implicit_reward}
    r_{\rm dpo}(x,y)=\beta\log\frac{\pi_{\rm dpo}(y|x)}{\pi_{\rm ref}(y|x)},  
\end{equation}
and token-level reward \cite{rafailov2024fromrtoq} can be further represented by 
\begin{equation} \label{eq:token_dpo_reward}
    r_{\rm dpo}(x,y_{<t},y_t)=\beta\log\frac{\pi_{\rm dpo}(y_t|y_{<t},x)}{\pi_{\rm ref}(y_t|y_{<t},x)}.
\end{equation}
These concise forms of reward further facilitate researches on self-rewarding \cite{chen2024bootstrapping} and fine-grained optimization \cite{xia2024inverseq*,zhong2024rto,yang2024sepo}.
Likewise, in this work, we also leverage the DPO reward and derive a RLHF-equivalent distillation objective for token-level reward optimization.

\section{Theoretical Analysis: From RLHF to Policy Distillation} \label{sec:theo_anal}
In this section, we provide a theoretical analysis for RLHF with DPO reward, building a connection between the objectives of RLHF and distillation.
%
As presented in Sec. \ref{sec:background_dpo}, DPO parameterizes the reward with the log ratio between two language models and trains it with the same objective of reward modeling.
Thus, the first intuition of this work is to substitute the reward in Eq. (\ref{eq:dpo_implicit_reward}) trained by DPO into the original RLHF objective:
\begin{align} 
    &\widetilde{\mathcal{J}}_{\rm RLHF}(\theta)= \nonumber \\
    &\max_{\theta}\!\!\!\!\mathop{\mathbb{E}}_{\substack{x \sim D \\ y \sim \pi_{\theta}(\cdot|x)}} \Big[ r_{\rm dpo}(x,y) - \beta \log \frac{\pi_{\theta}(y|x)}{\pi_{\rm ref}(y|x)} \Big] \label{eq:rlhf_with_dpo_reward1} \\
    &=\max_{\theta}\!\!\!\!\! \nonumber \\
    &\mathop{\mathbb{E}}_{\substack{x \sim D \\ y \sim \pi_{\theta}(\cdot|x)}} \!\!\! \Big[ \underbrace{\beta_0 \log \frac{\pi_{\rm dpo}(y|x)}{\pi_{\rm ref}(y|x)}}_{\rm DPO\ reward} - \beta \underbrace{\log \frac{\pi_{\theta}(y|x)}{\pi_{\rm ref}(y|x)}}_{\rm KL\ divergence} \Big], \label{eq:rlhf_with_dpo_reward2} 
\end{align}
where $\beta_0$ denotes the original coefficient in DPO training and is a constant in this objective.

It can be found that both the DPO reward and the KL divergence in Eq. (\ref{eq:rlhf_with_dpo_reward2}) can be decomposed into the sum of token-level results, which offers the potential to reformulate this objective into a token-level form.
Through solving this, we build a connection between RLHF with DPO reward and a policy distillation process, as described in the following theorem:

%


\begin{theorem} \label{thm:thm1}
    Under the DPO reward, the RLHF objective is equivalent to the following token-level policy distillation objective:
    \begin{align}
        &\max_{\theta}\!\!\!\! \mathop{\mathbb{E}}_{\substack{x \sim D \\ y \sim \pi_{\theta}(\cdot|x)}} \Big[ r_{\rm dpo} (x,y) - \beta \log \frac{\pi_{\theta}(y|x)}{\pi_{\rm ref}(y|x)} \Big] \label{eq:rlhf_with_dpo_reward}\\
        &= \min_{\theta} \mathop{\mathbb{E}}_{\substack{x \sim D \\ y \sim \pi_{\theta}(\cdot|x)}} \beta  \sum_{t=1}^{|y|} \nonumber \\
        &\hphantom{=\min_{\theta} {}} \mathbb{D}_{\rm KL}(\pi_{\theta}(\cdot|y_{<t},x)||\pi^{*}(\cdot|y_{<t},x)), \label{eq:aligndistil_obj_main}
    \end{align}
    where $\mathbb{D}_{\rm KL}(\cdot||\cdot)$ is token-level KL divergence and $\pi^{*}(\cdot|x,y_{<t})$ is the probability distribution output by the softmax function on a synthetic logit distribution $z^{*}_t$:
    \begin{equation} \label{eq:logit_distribution_main}
        z^{*}_t=\frac{\beta_0}{\beta}z^{\rm dpo}_{t} + (1 - \frac{\beta_0}{\beta})z^{\rm ref}_{t},
    \end{equation}
    where $z^{\rm dpo}_{t}$ and $z^{\rm ref}_{t}$ denote logit distributions of the DPO model and the reference model at $t$-th token position.
\end{theorem}
The proof is provided in Appendix \ref{sec:appendix_proof}.
Theorem \ref{thm:thm1} indicates that with DPO reward, we can equivalently convert the original sequence-level RLHF objective into a token-level distillation objective, thereby naturally achieving token-level reward optimization.

\section{AlignDistil}
In this section, we will introduce our AlignDistil motivated by the above theoretical analysis. 
Built on the theory, AlignDistil additionally introduces two intuitive designs, \emph{i.e.}, contrastive DPO reward (\S\ref{sec:ctr_dpo_reward}) and token adaptive logit extrapolation (\S\ref{sec:adaptive_logit}).
Lastly, we will conclude the objectives of AlignDistil for both on-policy and off-policy training (\S\ref{sec:overall_obj}).

\subsection{Contrastive DPO Reward} 
\label{sec:ctr_dpo_reward}
Although the DPO reward can theoretically represent any reward under the Bradley-Terry model \citep{rafailov2024dpo}, it has been pointed out to be less accurate than a pure reward model in practice \citep{lin2024limitdporeward}. 
We also observe this phenomenon in our experiments (see Table \ref{tab:reward_acc}) and conjecture that this imperfect reward estimation will impact the final alignment performance.
Thus, in AlignDistil, we parameterize the DPO reward by a pair of contrastive DPO models \citep{liu2024tisdpo}, \emph{i.e.}, a normal DPO model and a reverse DPO model (trained by switch chosen-rejected pairs in training data).
Intuitively, a reverse DPO model is more appropriate for the DPO reward as it captures negative features in low-quality data and makes the reward more discriminative.
Formally, this contrastive DPO reward can be represented as:
\begin{equation}
    r_{\rm ctr}(x,y)=\beta_0 \log \frac{\pi_{\rm dpo}(y|x)}{\pi^{-}_{\rm dpo}(y|x)},
\end{equation}
where $\pi^{-}_{\rm dpo}$ represents the reverse DPO model.
Note that the contrastive DPO reward introduces a new model $\pi^{-}_{\rm dpo}$ to the objective and increases the training cost.
To solve this, we switch the reference model in the RLHF objective from the initial model to the DPO model $\pi_{\rm dpo}$.
This not only saves the required models in training, but also moves the reference model forward for better alignment.
Afterward, the objective of RLHF in Eq. (\ref{eq:rlhf_with_dpo_reward}) becomes:
\begin{align} \label{eq:mod_rlhf}
    \max_{\theta}\!\!\!\! \mathop{\mathbb{E}}_{\substack{x \sim D \\ y \sim \pi_{\theta}(\cdot|x)}} \!\!\!\! \Big[ \beta_0 \log \frac{\pi_{\rm dpo}(y|x)}{\pi^{-}_{\rm dpo}(y|x)} - \beta \log \frac{\pi_{\theta}(y|x)}{\pi_{\rm dpo}(y|x)} \Big].
\end{align}
Correspondingly, the synthetic logit distribution in Eq. (\ref{eq:logit_distribution_main}) also changes to
\begin{align}
    z^{*}_t&=(1 + \frac{\beta_0}{\beta})z^{\rm dpo}_{t} - \frac{\beta_0}{\beta}z^{\rm dpo^{-}}_{t} \\
    &= \underbrace{z^{\rm dpo}_{t}}_{\rm DPO\ distribution} + \underbrace{\frac{\beta_0}{\beta} (z^{\rm dpo}_{t} - z^{\rm dpo^{-}}_{t})}_{\rm reward\ distribution}.
    \label{eq:logit_extrapolation}
\end{align}
The detailed derivation can be referred to in Appendix \ref{sec:logit_dist_change}.
Given that $\beta_0 > 0$ and $\beta > 0$, this equation strictly describes an extrapolation between logit distributions of the DPO model and the reverse DPO model.
The extrapolation is crucial for pushing the current policy to surpass the DPO model, since it constructs a stronger aligned distribution by removing some ``\textit{negative}'' information from the reverse DPO model and has been proven effective in \citep{liu2024dera}.
%

\subsection{Token Adaptive Logit Extrapolation}
\label{sec:adaptive_logit}
Although logit extrapolation theoretically provides a stronger distribution, we find that it is tricky to select an appropriate $\beta$ in practice.
Specifically, a large $\beta$ yields a small $\frac{\beta_0}{\beta}$ and may result in under-optimization, while a small $\beta$ produces a drastic distribution and tends to over-optimize the current policy.
Considering that tokens in the sequence have different tendencies, we design a token-level adaptive weight to adjust $\frac{\beta_0}{\beta}$ for each token position.
Specifically, we use the total variation distance (TVD)\footnote{We choose TVD since it is symmetric and computationally efficient with a limited range in $[0,1]$.} \citep{levin2017markov} between the DPO distribution and the reverse DPO distribution to calculate a coefficient $\alpha_t$ for position $t$:
\begin{equation}
    \alpha_t = \mathbb{D}_{\rm TVD}(t) * r + \epsilon \in [\epsilon, r+\epsilon],
\end{equation}
where $\mathbb{D}_{\rm TVD}(t) := \frac{1}{2} \sum_{y_t \in \mathcal{V}} | \pi_{\rm dpo}(y_t|y_{<t},x) - \pi^{-}_{\rm dpo}(y_t|y_{<t},x) |$, $\mathcal{V}$ is the vocabulary, $r$ is a hyper-parameter to control the upper-bound of the coefficient, and $\epsilon=0.001$ is a small value to avoid $\alpha_t=0$.
The intuition is that when DPO distribution is far from the reverse one, this position may have a key impact on the final reward and thus should learn from a stronger teacher distribution.
Accordingly, we modify Eq. (\ref{eq:logit_extrapolation}) as follows:

\begin{equation} \label{eq:ada_weight}
    z^{*}_t= z^{\rm dpo}_{t} + \alpha_t (z^{\rm dpo}_{t} - z^{\rm dpo^{-}}_{t}).
\end{equation}
Note that we replace the constant $\frac{\beta_0}{\beta}$ with an adaptive weight $\alpha_t$, and thus the static $\beta$ in Eq. (\ref{eq:aligndistil_obj_main}) also becomes adaptive as $\beta_t = \frac{\beta_0}{\alpha_t}$.
%
%
%

\subsection{Overall Objectives}
\label{sec:overall_obj}
The theoretical objective of AlignDistil follows Eq. (\ref{eq:aligndistil_obj_main}) with a synthetic teacher distribution $\pi^{*}$ calculated from Eq. (\ref{eq:ada_weight}).
It defines AlignDistil as an on-policy algorithm.
Practically, the loss for on-policy training of AlignDistil relies on Monte-Carlo sampling to estimate the expectation in Eq. (\ref{eq:aligndistil_obj_main}) and calculate the token-level distillation loss:
\begin{align} \label{eq:on_policy_obj}
    &\mathcal{L}_{\rm AD}^{\rm on} = \\
    &\frac{1}{|\mathcal{B}|} \sum_{x \in \mathcal{B}} \frac{\beta_t}{|\hat{y}|} \sum_{t=1}^{|\hat{y}|} \mathbb{D}_{\rm KL}( \pi_{\theta}(\cdot|\hat{y}_{<t},x)||\pi^{*}(\cdot|\hat{y}_{<t},x)), \nonumber
\end{align}
where $\mathcal{B}$ represents the mini-batch of prompts sampled from the prompt dataset $\{ (x)_i \}_{N}$ and $\hat{y}$ is sampled from $\pi_{\theta}(\cdot|x)$.
As this loss function is actually a supervised distillation loss, we can also construct an off-policy version using a prompt-response dataset $\{ (x,y)_i \}_{N}$:
\begin{align} \label{eq:off_policy_obj}
    &\mathcal{L}_{\rm AD}^{\rm off} = \\ 
    &\frac{1}{|\mathcal{B}|} \!\! \sum_{(x,y) \in \mathcal{B}} \! \frac{\beta_t}{|y|} \!\sum_{t=1}^{|y|} \mathbb{D}_{\rm KL}( \pi_{\theta}(\cdot|y_{<t},x)||\pi^{*}(\cdot|y_{<t},x)). \nonumber
\end{align}

\section{Experiments}
In this section, we present the experimental setups and showcase the evaluation results of our method.

\subsection{Experimental Setup}
\paragraph{Models.} In our experiments, we use two instruct models, \emph{i.e.}, \texttt{Qwen2-1.5B-Instruct} \citep{yang2024qwen2} and \texttt{Qwen2.5-1.5B-Instruct} \citep{yang2024qwen2_5} as the initial models for further alignment.
\paragraph{Datasets and Training.} Following most previous work \citep{meng2024simpo}, we use UltraFeedback \citep{cui2023ultrafeedback} as the training dataset, which contains about 63K prompts and corresponding response pairs with preference annotation.
Specifically, for DPO and reward modeling, we use both the prompts and the response pairs for training, while for other on-policy methods including ours, we only use the prompts for training.
For off-policy AlignDistil, we use the prompts and the preferred response in UltraFeedback.
For all experiments, we train the initial model for 1 epoch, with a batch size of 128, a learning rate of 1e-6, and a warmup ratio of 0.1.
All our experiments are conducted on 8 $\times$ A100-40G GPUs.
More training details are provided in Appendix \ref{sec:imple_detail}.
\paragraph{Evaluation.} Following the common practice \citep{meng2024simpo,kim2024spa}, we choose the following three benchmarks to evaluate the alignment performance of all the models: 
\begin{itemize}
    \item \textbf{AlpacaEval 2.0} \citep{dubois2024alpacaeval} consists of 805 instructions with the responses of GPT-4 as the baseline. 
    The evaluated responses are compared with the baseline by an LLM evaluator.
    We report the win rate (\textbf{WR}) and the length-controlled win rate (\textbf{LC WR}) for each model, where the LC WR is designed to eliminate the length bias in LLM-as-Judge.
    \item \textbf{MT-Bench} \citep{zheng2023mtbench} contains 80 multi-turn questions and assesses the quality of responses with scores between $[1, 10]$ by an LLM evaluator. We report the scores in the 1st turn (\textbf{1st Turn}) and the 2nd turn (\textbf{2nd Turn}) and the final averaged scores (\textbf{Avg.}).
    \item \textbf{Arena-Hard} \citep{li2024arenahard} incorporates 500 technical problem-solving queries with the responses of GPT-4 as the baseline.
    We report the win rate (\textbf{WR}) and the style-controlled win rate (\textbf{SC WR}) to mitigate the style bias in LLM evaluation.
\end{itemize}
We choose \texttt{Qwen2.5-72b-Instruct} as the automatic evaluator since we find that it achieves comparable judgment performance with GPT-4 with a much lower price (see Table \ref{tab:evaluators}).

\begin{table*}[t]
    \centering
    \resizebox{\linewidth}{!}{
    \begin{tabular}{l|cc|ccc|cc}
        \toprule
        \multirow{2}{*}{\textbf{{Methods}}} & \multicolumn{2}{c|}{\textbf{{AlpacaEval 2.0}}} & \multicolumn{3}{c|}{\textbf{{MT-Bench}}} & \multicolumn{2}{c}{\textbf{{Arena-Hard}}} \\
        \cmidrule(lr){2-3}\cmidrule(lr){4-6}\cmidrule(lr){7-8}
        & LC WR (\%) & WR (\%) & 1st Turn & 2nd Turn & Avg. & WR (\%) & SC WR (\%) \\
        \midrule
        \multicolumn{8}{c}{\textbf{\textit{Qwen2-1.5B-Instruct}}} \\
        \midrule
        Initial Model & 3.10 & 1.99 & 6.11 & 5.15 & 5.63 & 1.8 & 2.8 \\
        DPO \cite{rafailov2024dpo} & 6.42 & 5.03 & 6.19 & 5.59 & 5.89 & 3.0 & 3.6 \\
        DPO$_{\beta=0.01}$ \cite{rafailov2024dpo} & 10.72 & 11.61 & 6.70 & 6.06 & 6.38 & 7.0 & \textbf{6.8} \\
        KTO \citep{ethayarajh2024kto} & 7.16 & 6.34 & 6.54 & 5.55 & 6.05 & 3.4 & 4.2 \\
        SimPO \cite{meng2024simpo} & 8.19 & 9.63 & 5.94 & 5.71 & 5.83 & 6.9 & 4.2 \\
        TDPO$_{1}$ \cite{zeng2024tdpo} & 6.58 & 4.60 & 6.53 & 5.64 & 6.08 & 3.2 & 3.9 \\
        TDPO$_{2}$ \cite{zeng2024tdpo} & 3.59 & 2.42 & 6.25 & 5.06 & 5.66 & 1.3 & 1.9 \\
        PPO \cite{schulman2017ppo} & 4.86 & 4.41 & 6.76 & 5.51 & 6.13 & 2.7 & 3.0 \\
        RTO \cite{zhong2024rto} & 8.92 & 9.32 & 6.46 & 6.06 & 6.26 & 6.7 & 5.9 \\
        \midrule
        Off-Policy AlignDistil (ours) & 11.79 & 14.29 & 6.83 & 5.68 & 6.25 & 10.5 & 6.0 \\
        On-Policy AlignDistil (ours) & \textbf{12.93} & \textbf{15.65} & \textbf{6.89} & \textbf{6.13} & \textbf{6.45} & \textbf{11.0} & 6.7 \\
        \midrule
        \multicolumn{8}{c}{\textbf{\textit{Qwen2.5-1.5B-Instruct}}}  \\
        \midrule
        Initial Model & 12.57 & 8.94 & 7.15 & 6.05 & 6.60 & 16.8 & 12.7 \\
        DPO \cite{rafailov2024dpo} & 14.35 & 10.74 & 7.39 & 6.58 & 6.98 & 17.1 & 14.8 \\
        DPO$_{\beta=0.01}$ \cite{rafailov2024dpo} & 14.09 & 14.29 & 7.36 & 6.54 & 6.95 & 16.2 & 15.5 \\
        KTO \cite{ethayarajh2024kto} & 14.07 & 10.00 & 7.41 & 6.59 & 7.00 & 15.0 & 12.3 \\
        SimPO \cite{meng2024simpo} & 11.61 & 9.81 & 7.43 & 6.96 & 7.20 & 4.0 & 4.0 \\
        TDPO$_{1}$ \cite{zeng2024tdpo} & 13.19 & 9.94 & 7.45 & 6.66 & 7.06 & 16.5 & 14.1 \\
        TDPO$_{2}$ \cite{zeng2024tdpo} & 13.64 & 9.07 & 7.57 & 6.48 & 7.02 & 15.8 & 13.0 \\
        PPO \cite{schulman2017ppo} & 18.06 & 12.67 & 7.60 & 6.81 & 7.21 & 15.9 & 13.7 \\
        RTO \cite{zhong2024rto} & 16.54 & 15.53 & 7.37 & 6.51 & 6.94 & 18.2 & 16.6 \\
        \midrule
        Off-Policy AlignDistil (ours) & \textbf{21.16} & \textbf{24.29} & 7.62 & 6.53 & 7.07 & \textbf{24.1} & 21.8 \\
        On-Policy AlignDistil (ours) & 19.45 & 22.11 & \textbf{7.65} & \textbf{6.98} & \textbf{7.31} & 24.0 & \textbf{23.0} \\
        \bottomrule
    \end{tabular}}
    \caption{Evaluation results of baselines and our AlignDistil on three benchmarks. The best results are \textbf{bolded}. ``DPO$_{\beta=0.01}$'' represents DPO training with $\beta=0.01$.}
    \label{tab:main}
\end{table*}

\subsection{Baseline Methods}
We compare our method to the following methods:
\paragraph{DPO.} DPO \cite{rafailov2024dpo} is the most common direct preference learning method. 
The model trained by DPO is used both as a baseline and to calculate rewards for our method.
\paragraph{KTO.} KTO \citep{ethayarajh2024kto} is a direct preference learning method like DPO but optimizes on the non-paired preference data.
\paragraph{TDPO.} \citet{zeng2024tdpo} propose token-level DPO (TDPO) by equipping DPO reward with token-level forward KL constraint.
This method contains two versions, \emph{i.e.}, TDPO$_{1}$ and TDPO$_{2}$. 
\paragraph{SimPO.} SimPO \cite{meng2024simpo} is also a direct preference learning method and further simplifies DPO by removing the reference model.
\paragraph{PPO.} PPO \cite{schulman2017ppo} is selected as the default RL algorithm for RLHF, which optimizes the advantages of the policy estimated by generalized advantage estimator (GAE).
\paragraph{RTO.} \citet{zhong2024rto} propose reinforced token optimization (RTO) by substituting token-level DPO reward from Eq. (\ref{eq:token_dpo_reward}) into PPO.

The implementation details for these baselines are provided in Appendix \ref{sec:imple_detail}. 

\subsection{Main Results}
The evaluation results on three benchmarks are listed in Table \ref{tab:main}.
We can draw several conclusions from the results:
\textbf{1) Overall, both on-policy and off-policy AlignDistil significantly outperform baseline methods.}
%
%
Although the teacher distributions in our AlignDistil are constructed from DPO models, the performance of AlignDistil surpasses DPO by a notable margin (\emph{e.g.}, over 6 \% improvement for length-controlled win rates on AlpacaEval 2.0).
since \citet{liu2024dera} reveal that logit extrapolation in inference is similar to rescale $\beta$ in DPO training, we also implement another DPO with $\beta=0.01$ (noted as DPO$_{\beta=0.01}$).
We observe that rescaling $\beta$ does not always lead to improvement (\emph{e.g.}, the results on Qwen2.5-1.5B-Instruct), which indicates that a simple logit extrapolation may not stably improve the performance and the design of the contrastive DPO reward and token adaptive logit extrapolation are necessary.
\textbf{2) AlignDistil yields better token-level LLM alignment.}
TDPO$_{1/2}$ introduces token-level KL constraint for DPO, while this constraint may limit the performance on small models.
Besides, RTO introduces token-level DPO rewards into PPO and exhibits strong performance, especially surpassing PPO significantly on Qwen2-1.5B-Instruct.
This superiority highlights the benefits of token-level rewards in LLM alignment.
Nevertheless, our AlignDistil performs even better than RTO on both models since we further leverage the whole reward distribution instead of the scalar reward on the predicted token for optimization.
\textbf{3) Off-policy AlignDistil performs competitively to the on-policy version.}
Different from most methods for LLM alignment, our AlignDistil can work under both on-policy training and off-policy training.
On Qwen2-1.5B-Instruct, on-policy AlignDistil performs better, while off-policy AlignDistil performs comparably with the on-policy one on Qwen2.5-1.5B-Instruct.
We conjecture that for the DPO reward, the data for off-policy AlignDistil is in-distribution, while the data for on-policy AlignDistil is generated by the current policy model and is out-of-distribution.
These promising results suggest the off-policy AlignDistil as an efficient and effective method for LLM alignment.

%
%
%
%
%
%
%

\section{Analysis}
In this section, we first conduct the ablation study by separately analyzing the two designs in AlignDistil, \emph{i.e.}, the contrastive DPO reward (\S\ref{sec:ctr_dpo_reward_ablation}) and the token adaptive logit extrapolation (\S\ref{sec:adaptive_logit_ablation}).
Then we showcase the advantage of our AlignDistil on the convergence speed (\S\ref{sec:convergence}).

\subsection{DPO Reward: Contrastive vs. Vanilla}
\label{sec:ctr_dpo_reward_ablation}
\citet{lin2024limitdporeward} reveal that DPO reward often shows inferior generalization performance than a pure reward model.
We also verify this phenomenon in Table \ref{tab:reward_acc}.
Specifically, we calculate the response-level accuracy of different types of rewards on the training set and the test set of UltraFeedback.
Table \ref{tab:reward_acc} shows a performance gap between the DPO reward and the reward model.
By contrast, the contrastive DPO reward not only shows better accuracy than the vanilla DPO reward on training data, but also generalizes better on test data, even surpassing the reward model.
Correspondingly, our on-policy AlignDistil with the contrastive DPO reward outperforms the one with the vanilla DPO reward on AlpacaEval 2.0.
This performance gain can be attributed to the reverse DPO model, which captures subtle features in low-quality responses and implicitly \textbf{doubles the trainable parameters} in the DPO reward.
Therefore, the contrastive DPO reward plays a key role in our AlignDistil.

\begin{table}[t]
    \centering
    \resizebox{\linewidth}{!}{
    \begin{tabular}{l|c|c|c}
        \toprule
        \textbf{Reward} & \makecell{\textbf{Train} \\ Acc. (\%)} & \makecell{\textbf{Test} \\ Acc. (\%)} & \makecell{\textbf{AE2} \\ LC. (\%)} \\
        \midrule
        Reward Model & 70.41 & 71.19 & - \\
        DPO Reward & 72.85 & 69.53 & 16.51 \\
        Contrastive DPO Reward & \textbf{74.71} & \textbf{71.29} & \textbf{19.45} \\
        \bottomrule
    \end{tabular}
    }
    \caption{Reward accuracy of different types of rewards on 1000 samples from the training set and test set of UltraFeedback and the corresponding length-controlled win rate on AlpacaEval 2.0. All models are based on Qwen2.5-1.5B-Instruct.}
    \label{tab:reward_acc}
\end{table}

\subsection{Effect of Adaptive Logit Extrapolation}
\label{sec:adaptive_logit_ablation}
In our AlignDistil, we design a token adaptive logit extrapolation before constructing the teacher distribution.
The motivation is that a constant extrapolation weight $\frac{\beta_0}{\beta}$ for all tokens tends to over-optimize or under-optimize on some tokens.
Therefore, we explore whether this motivation holds.
Specifically, we set the extrapolation weight $\frac{\beta_0}{\beta}$ to a constant and test the performance and the KL divergence from the DPO model $\pi_{\rm dpo}$ as well as the average response length under different constants.
As shown in Table \ref{tab:token_adaptive}, when the constant is small (\emph{e.g.}, 1.0 and 1.2), the teacher distribution is similar to the distribution of the DPO model $\pi_{\rm dpo}$, reflecting by a small KL divergence.
However, the mild extrapolation also limits the strength of the teacher distribution and leads to under-optimization of the current policy.
By contrast, although a large extrapolation weight (\emph{e.g.}, 1.8 and 2.0) indeed yields better performance on AlpacaEval, the current policy is over-optimized, showcasing too much deviation from the DPO model and extremely increasing the response length.
Compared to these constant values, our token adaptive extrapolation weight considers the individual characteristics of different tokens and assigns an appropriate weight for each position, thus achieving a balance between performance and deviation.
\begin{table}[t]
    \centering
    \resizebox{\linewidth}{!}{
    \begin{tabular}{c|c|c|c|c}
        \toprule
        $\frac{\beta_0}{\beta}$ & \textbf{Type} & $\mathbb{D}_{\rm KL}(\pi_{\theta}||\pi_{\rm dpo})$ $\downarrow$ & \textbf{Avg. Length} $\downarrow$ & \makecell{\textbf{AE2} \\  LC. (\%) $\uparrow$} \\
        \midrule
        $1.0$ & constant & 10.16 & 2332 & 18.40 \\
        $1.2$ & constant & 13.95 & 2481 & 18.33 \\
        $1.5$ & constant & 20.71 & 2973 & 19.49 \\
        $1.8$ & constant & 28.55 & 3644 & 21.87 \\
        $2.0$ & constant & 34.52 & 4434 & 25.17 \\
        \midrule
        $\alpha_t$ & token adaptive & 22.95 & 2424 & 21.16 \\
        \bottomrule
    \end{tabular}}
    \caption{Comparisons between constant extrapolation weight and token adaptive extrapolation weight. The training is off-policy to mitigate the impact of data.}
    \label{tab:token_adaptive}
\end{table}

\begin{table}[t]
    \centering
    \resizebox{\linewidth}{!}{
    \begin{tabular}{l|c|c}
    \toprule
        \textbf{Methods} & \textbf{Win Rates (\%)} & \textbf{Avg. Reward} \\
        \midrule
        SFT & 50\% & 0.55 \\
        DPO & 85.7\% & 1.67 \\
        PPO & 87.7\% & 2.22 \\
        RTO & 90.8\% & 1.98 \\
        \midrule
        Off-Policy AlignDistil & \textbf{92.8\%} & 2.28 \\
        On-Policy AlignDistil & 92.5\% & \textbf{2.46} \\
    \bottomrule
    \end{tabular}}
    \caption{Win Rates (\%) compared to SFT baseline and the average reward scores of different methods on TL;DR.}
    \label{tab:tldr}
\end{table}

\subsection{Results on Larger Models}
To evaluate the effectiveness of our method on larger models, we also conduct experiments on Qwen2.5-7B and LLama3-8B.
For Qwen2.5-7B, we first train an instruction model following \citet{meng2024simpo} on UltraChat-200k \cite{ding2023ultrachat200k}.
For Llama3-8B, we directly use the SFT checkpoint open-sourced by \citet{meng2024simpo}.
As shown in Table \ref{tab:larger_models}, for both models, our AlignDistil exhibits significant improvement compared to existing baseline methods on AlpacaEval.
This promising performance sufficiently demonstrates the effectiveness of our method.

\begin{table}[t]
    \centering
    \resizebox{\linewidth}{!}{
    \begin{tabular}{l|cc}
        \toprule
        \multirow{2}{*}{\textbf{{Methods}}} & \multicolumn{2}{c}{\textbf{{AlpacaEval 2.0}}} \\
        \cmidrule(lr){2-3}
        & LC WR (\%) & WR (\%) \\
        \midrule
        \multicolumn{3}{c}{\textbf{\textit{Qwen2.5-7B}}} \\
        \midrule
        SFT & 7.51 & 3.66 \\
        DPO & 22.25 & 17.10 \\
        SimPO & 15.65 & 10.37 \\
        TDPO$_{1}$ & 26.42 & 19.44 \\
        TDPO$_{2}$ & 22.63 & 17.33 \\
        RTO & 17.75 & 12.00 \\
        \midrule
        Off-Policy AlignDistil \qquad\qquad & 29.16 & 23.82 \\
        On-Policy AlignDistil & \textbf{31.32} & \textbf{25.53} \\
        \midrule
        \multicolumn{3}{c}{\textbf{\textit{Llama3-8B}}}  \\
        \midrule
        SFT$^{\dagger}$ & 3.00 & 2.30 \\
        DPO$^{\dagger}$ & 15.67 & 15.58 \\
        SimPO$^{\dagger}$ & 11.24 & 12.92 \\
        TDPO$_{1}$ & 14.53 & 13.73 \\
        TDPO$_{2}$ & 14.48 & 13.79 \\
        RTO & 17.00 & 16.91 \\
        \midrule
        Off-Policy AlignDistil & 24.21 & \textbf{28.51} \\
        On-Policy AlignDistil & \textbf{26.29} & 27.64 \\
        \bottomrule
    \end{tabular}}
    \caption{Results of Qwen2.5-7B and Llama3-8B on AlpacaEval 2.0. Methods marked with ${\dagger}$ denote that the corresponding models are downloaded from the repository\protect\footnotemark of \citet{meng2024simpo} and directly evaluated under our setting.}
    \label{tab:larger_models}
\end{table}
\footnotetext{\url{https://huggingface.co/collections/princeton-nlp/simpo-66500741a5a066eb7d445889}}

\subsection{Performance on TL;DR}
To further evaluate the performance of our AlignDistil, we also conduct experiments using Qwen2.5-1.5B on TL;DR \cite{stiennon2022tldr}.
For evaluation, we calculate the win rate of each method compared to SFT with Qwen2.5-72B-Instruct as the judge model.
Besides, we calculate the average reward of each method by averaging the reward scores from the reward model in PPO training.
More details on training and evaluation are listed in Appendix \ref{sec:imple_detail_tldr}.
As shown in Table \ref{tab:tldr}, all alignment methods significantly outperform vanilla SFT.
On this basis, PPO further outperforms DPO on both win rates and reward, showcasing the benefits of RL training.
RTO shows a better win rate than PPO, while the reward is lower than that of PPO.
The reason we conjecture is that PPO directly optimizes this reward, while RTO optimizes the DPO reward.
Despite this, both our off- and on-policy AlignDistil surpass PPO and RTO on both metrics, which demonstrates the consistently exceptional performance of our method on different benchmarks.

\subsection{Convergence Speed}
\label{sec:convergence}
\begin{figure}
    \centering
    \includegraphics[width=\linewidth]{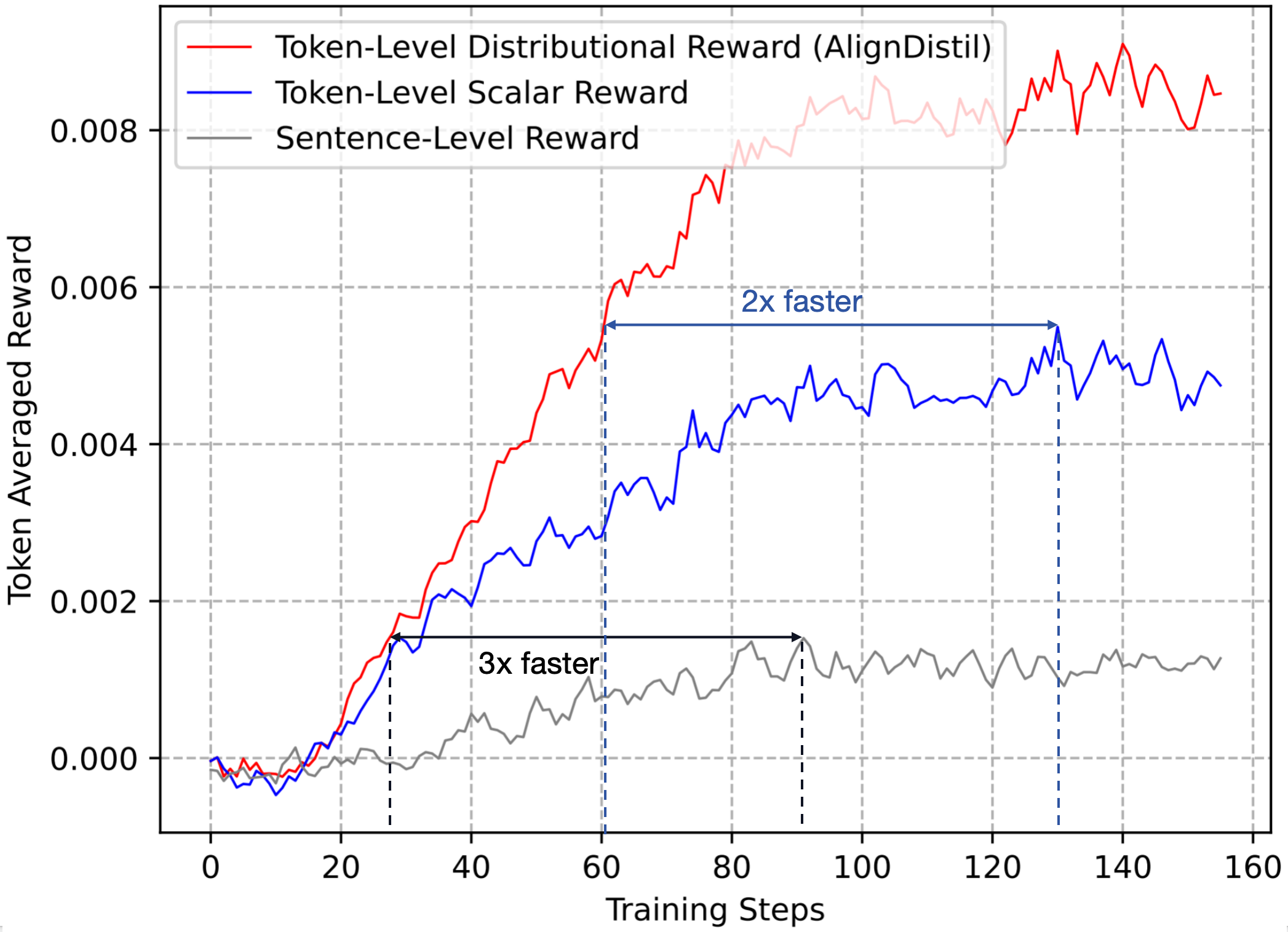}
    \caption{Convergence curves of token averaged reward from optimization on the sentence-level, token-level scalar-type, and token-level distributional reward.}
    \label{fig:convergence}
\end{figure}

As noted in previous literature \citep{chan2024densereward,zhong2024rto}, token-level reward optimization generally yields a faster convergence speed than sentence-level reward optimization.
We also test the convergence speed of our AlignDistil.
Specifically, we train the on-policy AlignDistil with a static $\beta$ on a subset (10k prompts) of UltraFeedback.
For comparison, we implement a sentence-level optimization method that optimizes the overall reward on the whole response as a bandit problem\footnote{The final reward involves the same contrastive DPO reward and KL constraint as AlignDistil.} and a REINFORCE-based method that optimizes on token-level scalar-type DPO rewards.
The coefficient $\beta$ is set to 0.08 for all methods\footnote{Implementation details are provided in Appendix \ref{sec:convergence_detail}.}.
The convergence curves on token averaged reward\footnote{The token averaged reward is used to mitigate the length bias in DPO reward.} corresponding to the training steps are plotted in Figure \ref{fig:convergence}.
It is shown that sentence-level reward optimization yields the poorest convergence, significantly lagging behind token-level rewards,
which is consistent with the conclusion in \citep{chan2024densereward,zhong2024rto}
Moreover, although token-level scalar reward boosts the convergence speed, our AlignDistil still has a more than 2$\times$ faster convergence speed.
The reason is that the optimization of AlignDistil leverages the whole reward distribution instead of a single scalar reward, allowing exact calculation of the reward expectation at each token position.
This comparison sufficiently demonstrates the benefits of our AlignDistil on token-level reward optimization.

\section{Related Work}
\paragraph{Fine-Grained LLM Alignment.}
Existing methods for LLM alignment are criticized for optimizing sparse, coarse-grained rewards.
To address this, \cite{lightman2023letsverifystepbystep} propose the process reward model (PRM) trained with step-level human annotations for complicated LLM reasoning.
On this basis, several methods are proposed to automatically collect step-level rewards without human annotation \citep{wang2024mathshepherd,luo2024improve,yuan2024freeprm}.
Besides, \citet{cao2024spanreward} extract span-level rewards from LLM critiques to enhance the PPO algorithm.
Furthermore, there are solutions for token-level reward signals via edit distance \citep{guo2024beyondimit,chen2024finegrained_editdis}, attention scores in the reward model \citep{chan2024densereward}, and reward model outputs on intermediate tokens \citep{li2024r3hf}.
Additionally, \citet{rafailov2024fromrtoq} reveal that DPO also automatically learns token-level reward.
Afterward, this token-level DPO reward is applied to existing alignment methods like DPO \citep{liu2024tisdpo,yang2024sepo} and PPO \citep{zhong2024rto} or new algorithms \citep{xia2024inverseq*}.
Following this line, we also leverage the DPO reward in our method, while the difference is that we further exploit the distributional information in this reward for more sufficient optimization.

\paragraph{Knowledge Distillation for LLMs.}
Knowledge distillation (KD, \citealp{hinton2015kd}) is proposed as an essential technique for compressing neural networks.
%
%
With the emergence and development of LLMs, KD has attracted more attention to reduce the numerous parameters in LLMs.
In this context, KD methods are divided into white-box KD \citep{hinton2015kd} and black-box KD \citep{kim2016seqkd}, depending on whether the weight of the teacher model can be obtained.
For white-box KD, approaches typically bridge probability distributions \cite{agarwal2024gkd,gu2024minillm,ko2024distillm,zhang2024dskd,zhang2025dskd_v2} or intermediate features \cite{wang2020minilm} between the teacher model and the student model.
Intuitively, this process transfers sufficient information from the teacher model, thus often used for pre-training small yet powerful LLMs \cite{team2024gemma,meta2024llama32}.
In contrast, black-box KD is actually more widely used for LLMs as it only requires collecting outputs from the teacher model and supervised fine-tuning the student model \cite{alpaca,vicuna2023,tunstall2023zephyr}.
Different from these methods, our AlignDistil is derived from RLHF and aims for token-level reward optimization.

\section{Conclusion}
In this paper, we aim at the fine-grained LLM alignment problem and propose AlignDistil as the solution.
Specifically, we introduce the DPO reward into the objective of RLHF and theoretically build an equivalence between RLHF and token-level policy distillation.
On this basis, we design two components in our AlignDistil, \emph{i.e.}, the contrastive DPO reward and token adaptive logit extrapolation, for better performance and stable optimization.
%
%
Experimental results on prevalent alignment benchmarks sufficiently demonstrate the superiority of our AlignDistil compared to existing methods for LLM alignment.
Moreover, we showcase that the token-level distributional reward optimization in AlignDistil offers a faster convergence speed than sentence-level and token-level scalar-type rewards.

\section*{Limitations}
Due to the resource limitation, the evaluation of our AlignDistil is limited to small language models (\textasciitilde 1.5B).
%
The evaluation of AlignDistil on larger models is still under-explored, and we leave this for future work.

\section*{Acknowledgments}
The research work described in this paper has been supported by the Fundamental Research Funds for the Central Universities (2024JBZY019) and the National Nature Science Foundation of China (No. 62476023, 61976016, 62376019), and the authors would like to thank the anonymous reviewers for their valuable comments and suggestions to improve this paper.

\bibliography{custom}

\onecolumn
\newpage
\appendix

\section{Proof of Theorem 1}
\label{sec:appendix_proof}
Here we recall Theorem \ref{thm:thm1}:
\begin{theorem*}
    Under the DPO reward, the RLHF objective is equivalent to the following token-level policy distillation objective:
    \begin{align}
        &\hphantom{= {}}\max_{\theta} \mathop{\mathbb{E}}_{\substack{x \sim D \\ y \sim \pi_{\theta}(\cdot|x)}} \Big[ r_{\rm dpo} (x,y) - \beta \log \frac{\pi_{\theta}(y|x)}{\pi_{\rm ref}(y|x)} \Big] \\
        &= \min_{\theta} \mathop{\mathbb{E}}_{\substack{x \sim D \\ y \sim \pi_{\theta}(\cdot|x)}} \beta  \sum_{t=1}^{|y|} \mathbb{D}_{\rm KL}(\pi_{\theta}(\cdot|x,y_{<t})||\pi^{*}(\cdot|x,y_{<t})), \label{eq:aligndistil_obj}
    \end{align}
    where $\mathbb{D}_{\rm KL}(\cdot||\cdot)$ is token-level KL divergence and $\pi^{*}(\cdot|x,y_{<t})$ is the probability distribution output by the softmax function on a synthetic logit distribution $z^{*}_t$:
    \begin{equation} \label{eq:logit_distribution}
        z^{*}_t=\frac{\beta_0}{\beta}z^{\rm dpo}_{t} + (1 - \frac{\beta_0}{\beta})z^{\rm ref}_{t},
    \end{equation}
    where $z^{\rm dpo}_{t}$ and $z^{\rm ref}_{t}$ denote logit distributions of the DPO model and the reference model at $t$-th token position.
\end{theorem*}

\begin{proof}
    First, we need to decompose the objective of RLHF into token level.
    Inspired by \citet{wen2023fdistill}, we derive the decomposition process from the objective in Eq. (\ref{eq:rlhf_with_dpo_reward2}):
    \begin{align}
        \widetilde{\mathcal{J}}_{\rm RLHF} (\theta) &= \max_{\theta} \mathop{\mathbb{E}}_{\substack{x \sim D \\ y \sim \pi_{\theta}(\cdot|x)}} \Big[ \beta_0 \log \frac{\pi_{\rm dpo}(y|x)}{\pi_{\rm ref}(y|x)} -\beta \log \frac{\pi_{\theta}(y|x)}{\pi_{\rm ref}(y|x)} \Big] \nonumber \\
        &= \max_{\theta} \mathop{\mathbb{E}}_{\substack{x \sim D \\ y_{1:T} \sim \pi_{\theta}(\cdot|x)}} \Big[ \beta_0 \sum_{t=1}^{T} \log \frac{\pi_{\rm dpo}(y_t|y_{<t},x)}{\pi_{\rm ref}(y_t|y_{<t},x)} -\beta \sum_{t=1}^{T} \log \frac{\pi_{\theta}(y_t|y_{<t},x)}{\pi_{\rm ref}(y_t|y_{<t},x)} \Big] \\
        &= \max_{\theta} \mathop{\mathbb{E}}_{\substack{x \sim D \\ y_{1:T} \sim \pi_{\theta}(\cdot|x)}} \sum_{t=1}^{T} \Big[ \beta_0 \log \frac{\pi_{\rm dpo}(y_t|y_{<t},x)}{\pi_{\rm ref}(y_t|y_{<t},x)} -\beta \log \frac{\pi_{\theta}(y_t|y_{<t},x)}{\pi_{\rm ref}(y_t|y_{<t},x)} \Big] \\
        &= \max_{\theta} \mathop{\mathbb{E}}_{\substack{x \sim D \\ y_{1:T} \sim \pi_{\theta}(\cdot|x)}} \sum_{t=1}^{T-1} \Big[ \beta_0 \log \frac{\pi_{\rm dpo}(y_t|y_{<t},x)}{\pi_{\rm ref}(y_t|y_{<t},x)} -\beta \log \frac{\pi_{\theta}(y_t|y_{<t},x)}{\pi_{\rm ref}(y_t|y_{<t},x)} \Big] \\
        &+ \mathop{\mathbb{E}}_{\substack{x \sim D \\ y_{1:T} \sim \pi_{\theta}(\cdot|x)}} \Big[ \beta_0 \log \frac{\pi_{\rm dpo}(y_T|y_{<T},x)}{\pi_{\rm ref}(y_T|y_{<T},x)} -\beta \log \frac{\pi_{\theta}(y_T|y_{<T},x)}{\pi_{\rm ref}(y_T|y_{<T},x)} \Big] \label{eq:last_step_expect} \\
        &= \max_{\theta} \mathop{\mathbb{E}}_{\substack{x \sim D \\ y_{1:T-1} \sim \pi_{\theta}(\cdot|x)}} \sum_{t=1}^{T-1} \Big[ \beta_0 \log \frac{\pi_{\rm dpo}(y_t|y_{<t},x)}{\pi_{\rm ref}(y_t|y_{<t},x)} -\beta \log \frac{\pi_{\theta}(y_t|y_{<t},x)}{\pi_{\rm ref}(y_t|y_{<t},x)} \Big] \\
        &+ \mathop{\mathbb{E}}_{\substack{x \sim D \\ y_{1:T-1} \sim \pi_{\theta}(\cdot|x)}} \sum_{y_t \in \mathcal{V}} \pi_{\theta}(y_t|y_{<T},x) \Big[ \beta_0 \log \frac{\pi_{\rm dpo}(y_t|y_{<T},x)}{\pi_{\rm ref}(y_t|y_{<T},x)} -\beta \log \frac{\pi_{\theta}(y_t|y_{<T},x)}{\pi_{\rm ref}(y_t|y_{<T},x)} \Big] \label{eq:decompose_last_step_expect}
    \end{align}
    Eq. (\ref{eq:decompose_last_step_expect}) is derived by decomposing the expectation in Eq. (\ref{eq:last_step_expect}) at the last step and exactly calculating it.
    Likewise, we can recursively decompose the expectation from step $T-1$ to step $1$ and obtain the final token-level representation:
    \begin{equation}
    \label{eq:token_rlhf_obj}
        \widetilde{\mathcal{J}}_{\rm RLHF} (\theta) = \max_{\theta} \mathop{\mathbb{E}}_{\substack{x \sim D \\ y \sim \pi_{\theta}(\cdot|x)}} \sum_{t=1}^{|y|} \sum_{y_t \in \mathcal{V}} \pi_{\theta}(y_t|y_{<t},x) \Big[ \beta_0 \log \frac{\pi_{\rm dpo}(y_t|y_{<t},x)}{\pi_{\rm ref}(y_t|y_{<t},x)} -\beta \log \frac{\pi_{\theta}(y_t|y_{<t},x)}{\pi_{\rm ref}(y_t|y_{<t},x)} \Big]
    \end{equation}
    Then, we reorganize the log ratio in Eq. (\ref{eq:token_rlhf_obj}):
    \begin{align}
        \widetilde{\mathcal{J}}_{\rm RLHF} (\theta) &= \max_{\theta} \mathop{\mathbb{E}}_{\substack{x \sim D \\ y \sim \pi_{\theta}(\cdot|x)}} \sum_{t=1}^{|y|} \sum_{y_t \in \mathcal{V}} \pi_{\theta}(y_t|y_{<t},x) \Big[ \beta_0 \log \frac{\pi_{\rm dpo}(y_t|y_{<t},x)}{\pi_{\rm ref}(y_t|y_{<t},x)} -\beta \log \frac{\pi_{\theta}(y_t|y_{<t},x)}{\pi_{\rm ref}(y_t|y_{<t},x)} \Big] \nonumber \\
        &=\max_{\theta} \mathop{\mathbb{E}}_{\substack{x \sim D \\ y \sim \pi_{\theta}(\cdot|x)}} \beta \sum_{t=1}^{|y|} \sum_{y_t \in \mathcal{V}} \pi_{\theta}(y_t|y_{<t},x) \Big[ \frac{\beta_0}{\beta} \log \pi_{\rm dpo}(y_t|y_{<t},x) \nonumber \\
        &\qquad\qquad\qquad\qquad\qquad + (1 - \frac{\beta_0}{\beta}) \log \pi_{\rm ref}(y_t|y_{<t},x) - \log \pi_{\theta}(y_t|y_{<t},x) \Big] \label{eq:decompose_log}
    \end{align}
    Here we introduce an equivalence between log probabilities and logits, \emph{i.e.}, when 
    \begin{align}
         p (i) = \frac{{\rm e}^{z_i}}{\sum_{j=1}^{|\mathcal{V}|} {\rm e}^{z_j}}, 
    \end{align}
    we have
    \begin{align} \label{eq:log_logits_rel}
        \log p (i) = z_i - \log \sum_{j=1}^{|\mathcal{V}|} {\rm e}^{z_j}.
    \end{align}
    Then, we substitute Eq. (\ref{eq:log_logits_rel}) into Eq. (\ref{eq:decompose_log}):
    \begin{align}
        \widetilde{\mathcal{J}}_{\rm RLHF} (\theta) &= \max_{\theta} \mathop{\mathbb{E}}_{\substack{x \sim D \\ y \sim \pi_{\theta}(\cdot|x)}} \beta \sum_{t=1}^{|y|} \sum_{y_t \in \mathcal{V}} \pi_{\theta}(y_t|y_{<t},x) \Big[ \frac{\beta_0}{\beta} \log \pi_{\rm dpo}(y_t|y_{<t},x) \nonumber \\
        &\qquad\qquad\qquad\qquad\qquad + (1 - \frac{\beta_0}{\beta}) \log \pi_{\rm ref}(y_t|y_{<t},x) - \log \pi_{\theta}(y_t|y_{<t},x) \Big] \\
        &= \max_{\theta} \mathop{\mathbb{E}}_{\substack{x \sim D \\ y \sim \pi_{\theta}(\cdot|x)}} \beta \sum_{t=1}^{|y|} \sum_{y_t \in \mathcal{V}} \pi_{\theta}(y_t|y_{<t},x) \Big[ \frac{\beta_0}{\beta} z^{\rm dpo}_t + (1 - \frac{\beta_0}{\beta}) z^{\rm ref}_t + Z - \log \pi_{\theta}(y_t|y_{<t},x) \Big] \label{eq:obj_logp_to_logits} \\
        &= \max_{\theta} \mathop{\mathbb{E}}_{\substack{x \sim D \\ y \sim \pi_{\theta}(\cdot|x)}} \beta \sum_{t=1}^{|y|} \sum_{y_t \in \mathcal{V}} \pi_{\theta}(y_t|y_{<t},x) \Big[ z^{*}_t - \log \pi_{\theta}(y_t|y_{<t},x) \Big],
    \end{align}
    where $z^{*}=\frac{\beta_0}{\beta} z^{\rm dpo}_t + (1 - \frac{\beta_0}{\beta}) z^{\rm ref}_t$, and $Z$ is a constant representing the logsumexp term in Eq. (\ref{eq:log_logits_rel}).
    Thus, it has no influence on the expectation and can be omitted in the later calculation.
    Then we leverage the equivalence again and convert the logits back to log probabilities:
    \begin{align}
        \widetilde{\mathcal{J}}_{\rm RLHF} (\theta) &= \max_{\theta} \mathop{\mathbb{E}}_{\substack{x \sim D \\ y \sim \pi_{\theta}(\cdot|x)}} \beta \sum_{t=1}^{|y|} \sum_{y_t \in \mathcal{V}} \pi_{\theta}(y_t|y_{<t},x) \Big[ z^{*}_t - \log \pi_{\theta}(y_t|y_{<t},x) \Big] \nonumber \\
        &= \max_{\theta} \mathop{\mathbb{E}}_{\substack{x \sim D \\ y \sim \pi_{\theta}(\cdot|x)}} \beta \sum_{t=1}^{|y|} \sum_{y_t \in \mathcal{V}} \pi_{\theta}(y_t|y_{<t},x) \Big[ \log \pi^{*}(y_t|y_{<t},x) - \log \pi_{\theta}(y_t|y_{<t},x) \Big] \\
        &= \max_{\theta} \mathop{\mathbb{E}}_{\substack{x \sim D \\ y \sim \pi_{\theta}(\cdot|x)}} \beta \sum_{t=1}^{|y|} \sum_{y_t \in \mathcal{V}} \pi_{\theta}(y_t|y_{<t},x)  \log \frac{ \pi^{*}(y_t|y_{<t},x) }{ \pi_{\theta}(y_t|y_{<t},x)} \\
        &= \min_{\theta} \mathop{\mathbb{E}}_{\substack{x \sim D \\ y \sim \pi_{\theta}(\cdot|x)}} \beta  \sum_{t=1}^{|y|} \sum_{y_t \in \mathcal{V}} \mathbb{D}_{\rm KL}(\pi_{\theta}(y_t|x,y_{<t})||\pi^{*}(y_t|x,y_{<t})).
    \end{align}
    Thus, the theorem is proved.
\end{proof}

\section{Derivations for Changed Logit Distribution}
\label{sec:logit_dist_change}
As shown in Eq. (\ref{eq:mod_rlhf}), we rewrite the objective of RLHF under the contrastive DPO reward as follows:
\begin{equation}
    \widetilde{\mathcal{J}}_{\rm RLHF}(\theta) = \max_{\theta} \mathop{\mathbb{E}}_{\substack{x \sim D \\ y \sim \pi_{\theta}(\cdot|x)}} \Big[ \beta_0 \log \frac{\pi_{\rm dpo}(y|x)}{\pi^{-}_{\rm dpo}(y|x)} - \beta \log \frac{\pi_{\theta}(y|x)}{\pi_{\rm dpo}(y|x)} \Big].
\end{equation}
Correspondingly, the token-level objective becomes
\begin{align}
    \label{eq:token_rlhf_obj}
        \widetilde{\mathcal{J}}_{\rm RLHF} (\theta) &= \max_{\theta} \mathop{\mathbb{E}}_{\substack{x \sim D \\ y \sim \pi_{\theta}(\cdot|x)}} \sum_{t=1}^{|y|} \sum_{y_t \in \mathcal{V}} \pi_{\theta}(y_t|y_{<t},x) \Big[ \beta_0 \log \frac{\pi_{\rm dpo}(y_t|y_{<t},x)}{\pi^{-}_{\rm dpo}(y_t|y_{<t},x)} -\beta \log \frac{\pi_{\theta}(y_t|y_{<t},x)}{\pi_{\rm dpo}(y_t|y_{<t},x)} \Big] \\
        &= \max_{\theta} \mathop{\mathbb{E}}_{\substack{x \sim D \\ y \sim \pi_{\theta}(\cdot|x)}} \beta \sum_{t=1}^{|y|} \sum_{y_t \in \mathcal{V}} \pi_{\theta}(y_t|y_{<t},x) \Big[ (1 + \frac{\beta_0}{\beta}) \log \pi_{\rm dpo}(y_t|y_{<t},x) \nonumber \\
        &\qquad\qquad\qquad\qquad\qquad\qquad - \frac{\beta_0}{\beta} \log \pi^{-}_{\rm dpo}(y_t|y_{<t},x) - \log \pi_{\theta}(y_t|y_{<t},x) \Big]. 
    \end{align}
The following derivation is similar to the one after Eq. (\ref{eq:decompose_log}) and thus omitted.

\section{Implementation Details}
\label{sec:imple_detail}
In this section, we provide details on the implementation of baseline methods and our AlignDistil.
All our implementation is based on the open-source toolkit OpenRLHF\footnote{\url{https://github.com/OpenRLHF/OpenRLHF}}.

\subsection{UltraFeedback} \label{sec:imple_detail_ufb}
Below, we list the individual settings for each method on UltraFeedback:
\begin{itemize}
    \item \textbf{DPO} (default setting): we set $\beta=0.1$ and optimize the model on UltraFeedback;
    \item \textbf{DPO$_{\beta=0.01}$}: the only difference compared to DPO (default setting) is $\beta=0.01$;
    \item \textbf{KTO}: we set $\beta=0.1$ and use the unpaired version of UltraFeedback for training;
    \item \textbf{TDPO$_1$}: similar to DPO in the default setting, we set $\beta=0.1$;
    \item \textbf{TDPO$_2$}: TDPO$_2$ introduces a new hyper-parameter $\alpha$ to control the intensity of KL term and we find that $\alpha=0.1$ works best in our experiments;
    \item \textbf{SimPO}: SimPO involves two hyper-parameters, \emph{i.e.}, $\beta$ and a ratio $\frac{\gamma}{\beta}$, which are needed fine-grained tuning to achieve ideal performance.
    Specifically, for both models, we set $\beta=10$ and $\frac{\gamma}{\beta}=0.5$ after extensive tuning.
    \item \textbf{PPO}: Before PPO, we first train a reward model on UltraFeedback based on Qwen2.5-1.5B-Instruct.
    We use the same reward model for both initial models since we find the reward model based on Qwen2-1.5B-Instruct leads to unstable PPO optimization.
    Afterward, we mainly follow the suggested settings in OpenRLHF for PPO training, \emph{e.g.}, setting the critic learning rate to 9e-6, rollout batch size to 1024, and the KL coefficient to 0.01.
    \item \textbf{RTO}: The procedure of RTO is similar to PPO, except for the token-level DPO reward $\beta \log \frac{\pi_{\rm dpo}(y_t|y_{<t},x)}{\pi_{\rm ref}(y_t|y_{<t},x)}$.
    We use the DPO model in the default setting with $\beta=0.1$ to calculate the DPO reward.
    Besides, RTO set $\beta_2$ as the KL coefficient in PPO.
    In our experiment, we find RTO is sensitive to $\beta_2$ and tends to produce overly long responses.
    Thus, we set $\beta_2=0.05$ as an appropriate value for stable training.
    After our experiments, the authors of RTO update their methods to fix this issue in the latest (v3) version of the paper \citep{zhong2024rto}.
    Despite better performance, this update is a concurrent work with ours and our implementation of RTO is still based on the v2 version of the paper.
    \item \textbf{On-Policy AlignDistil}: The on-policy AlignDistil uses the DPO model in the default setting as well as a reverse DPO model trained by switching the chosen/rejected responses in DPO training.
    For on-policy AlignDistil, we only use the prompts in Ultrafeedback and sample responses from the current policy.
    The hyper-parameter $r$ for token adaptive logit extrapolation is set to 20 for Qwen2-1.5B-Instruct and 15 for Qwen2.5-1.5B-Instruct.
    \item \textbf{Off-Policy AlignDistil}: For off-policy AlignDistil, we use both the prompts and the chosen responses in Ultrafeedback for training.
    The hyper-parameter $r$ is set to 10 for Qwen2-1.5B-Instruct and 15 for Qwen2.5-1.5B-Instruct.
    
\end{itemize}

\subsection{TL;DR} \label{sec:imple_detail_tldr}
Here, we provide the implementation details of the experiments on TL;DR.
\paragraph{Training.}
We start from the base model Qwen2.5-1.5B-Base and first conduct SFT on the SFT version of TL;DR\footnote{\url{https://huggingface.co/datasets/trl-lib/tldr}}.
Based on the checkpoint after SFT, we further apply LLM alignment algorithms, including DPO, PPO, RTO and our AlignDistil using the preference dataset of TL;DR\footnote{\url{https://huggingface.co/datasets/trl-lib/tldr-preference}}.
The configurations of all training procedures are listed below:
\begin{itemize}
    \item \textbf{SFT}: We train the Qwen2.5-1.5B model on the SFT dataset of TL;DR dataset for 1 epoch, with a batch size of 128 and a learning rate of 2e-5. \\
    \item \textbf{Reward Modeling}: Based on the checkpoint after SFT, we train a reward model on the preference data of TL;DR dataset for 1 epoch, with a batch size of 128 and a learning rate of 1e-6. This model is used for \textbf{both PPO training and evaluation}. \\
    \item \textbf{DPO}: In DPO training, $\beta$ is set to 0.1. The batch size and the learning rate are set to 128 and 1e-6 for DPO and all the following methods. \\
    \item \textbf{PPO}: The learning rate of the critic model is set to 9e-6 and the KL coefficient is set to 0.07. Other settings follow the default settings in OpenRLHF.
    \item \textbf{RTO}: $\beta$ is set to 0.1 and $\beta_2$ is set to 0.07. Other settings are kept the same with PPO.
    \item \textbf{Off-Policy AlignDistil}: $r$ is set to 1.
    \item \textbf{Off-Policy AlignDistil}: $r$ is set to 2.
\end{itemize}

\paragraph{Evaluation.}
During evaluation, we randomly sample 1000 items of data from the validation set of TL;DR-preference.
We use Qwen2.5-72B-Instruct to judge the win rates of different methods against SFT.
The prompt for judgement follows \cite{ahmadian2024rloo}.
Moreover, we leverage the reward model trained for PPO to measure the average reward of the models after different methods.

\section{Performance for Qwen2.5-72B-Instruct as Judge}
\label{sec:qwen_as_judge}
Qwen2.5-72B-Instruct has been demonstrated as a strong open-source model with comparable performance against the state-of-the-art LLMs.
Following the tools for evaluating LLM-as-Judge provided in the repository\footnote{\url{https://github.com/tatsu-lab/alpaca_eval}} of \citep{dubois2024alpacaeval}, we test the evaluation performance for Qwen2.5-72B-Instruct and list the performance in Table \ref{tab:evaluators}.
\begin{table}[h]
    \centering
    \resizebox{\linewidth}{!}{
    \begin{tabular}{l|cccc}
    \toprule
       Evaluators & Human Agreement & Price [\$/1000 examples] & Spearman corr. & Pearson corr. \\
       \midrule
        alpaca\_eval\_gpt4 & 69.17 & 13.60 & 0.97 & 0.93 \\
        alpaca\_eval\_gpt4\_turbo\_fn & 68.09 & 5.53 & 0.93 & 0.82 \\
        Qwen2.5-72B-Instruct & 67.63 & 0 & 0.92 & 0.86 \\
        weighted\_alpaca\_eval\_gpt4\_turbo & 65.73 & 4.32 & 0.78 & 0.77 \\
        humans & 65.66 & 300 & 1.00 & 1.00 \\
        \bottomrule
    \end{tabular}}
    \caption{Comparisons of Qwen2.5-72B-Instruct and some top evaluators on the AlpacaEval leaderboard in terms of performance and cost. We select several key columns from the leaderboard.}
    \label{tab:evaluators}
\end{table}

As shown in Table \ref{tab:evaluators}, Qwen2.5-72B-Instruct achieves comparable human agreement with \texttt{alpaca\_eval\_gpt4\_turbo\_fn} and \texttt{alpaca\_eval\_gpt4} with a much lower price since we can deploy the model with vLLM locally. 
Moreover, compared to the official recommended evaluator \texttt{weighted\_alpaca\_eval\_gpt4\_turbo}, Qwen2.5-72B-Instruct performs significantly better on both performance and cost.
Thus, we choose Qwen2.5-72B-Instruct as the evaluator for the three benchmarks.

\section{Implementation Details for Convergence Speed Comparison}
\label{sec:convergence_detail}
To evaluate the convergence speed of our AlignDistil, we use two methods that optimize sentence-level (response-level) rewards and token-level scalar-type rewards, respectively.
For sentence-level optimization, we use the contrastive DPO reward on the whole sequence and calculate the gradient of the policy model as follows:
\begin{equation}
    \nabla_{\theta} \mathcal{J}(\theta) = \frac{1}{|y|}\Big[ \sum_{t=1}^{|y|} r_{\rm ctr}(x,y) \nabla_{\theta} \log \pi_{\theta} (y_t|y_{<t},x) - \beta \nabla_{\theta} \log \frac{\pi_{\theta}(y|x)}{\pi_{\rm dpo}(y|x)} \Big].
\end{equation}
For token-level optimization with scalar-type rewards, we optimize token-level contrastive reward with a REINFORCE algorithm:
\begin{equation}
    \nabla_{\theta} \mathcal{J}(\theta) = \frac{1}{|y|} \sum_{t=1}^{|y|} \Big[  G_t \nabla_{\theta} \log \pi_{\theta} (y_t|y_{<t},x) - \beta \nabla_{\theta} \mathbb{D}_{\rm KL}(\pi_{\theta} (\cdot|y_{<t},x)||\pi_{\rm dpo} (\cdot|y_{<t},x)) \Big],
\end{equation}
where $G_t=\sum_{i=t}^{|y|} r_{\rm ctr}(x,y_{<i},y_i)$ is the return at position $t$.

\end{document}